%% file: main.tex
\documentclass[10pt,twocolumn,letterpaper]{article}

\pdfoutput=1

\usepackage{cvpr}
\usepackage{times}
\usepackage{epsfig}
\usepackage{graphicx}
\usepackage{amsmath}
\usepackage{amssymb}
\usepackage{color}
\usepackage{multirow}
\usepackage{tabularx}

\usepackage{cases}
\usepackage{wrapfig}
\usepackage{pdfpages}
\usepackage{sidecap}
\usepackage{colortbl}
\usepackage{overpic}
\usepackage{booktabs}
\usepackage{colortbl}
\usepackage[toc,page]{appendix}
\usepackage{placeins}

\usepackage{rotating}
\usepackage[export]{adjustbox}
\usepackage[table]{xcolor}
\usepackage[inline]{enumitem}

\definecolor{red}{rgb}{0.95,0.4,0.4}
\definecolor{blue}{rgb}{0.4,0.4,0.95}
\definecolor{darkblue}{rgb}{0,0,0.8}
\definecolor{darkred}{rgb}{0.8,0,0}
\definecolor{darkgreen}{rgb}{0,0.5,0}
\definecolor{grey}{rgb}{0.6,0.6,0.6}
\definecolor{col1}{RGB}{232, 161, 148}
\definecolor{col2}{RGB}{148, 187, 232}

\usepackage{xspace}

\newlength\savewidth\newcommand\shline{\noalign{\global\savewidth\arrayrulewidth
  \global\arrayrulewidth 1pt}\hline\noalign{\global\arrayrulewidth\savewidth}}

\graphicspath{{figs/matterport_results/}{figs/}{figs/kitti_results/}{figs/supp_figs/matterport_results/}}

\renewcommand*{\eg}{e.g.\@\xspace}

\usepackage{pifont}

\def\cvprPaperID{2582} 

\ifcvprfinal\fi

\cvprfinalcopy
\renewcommand{\paragraph}[1]{\noindent\textbf{#1}}

\begin{document}

\title{Footprints and Free Space from a Single Color Image}

\author{Jamie Watson$^1$
\hspace{20pt}
Michael Firman$^1$
\hspace{20pt}
Aron Monszpart$^1$
\hspace{20pt}
Gabriel J. Brostow$^{1, 2}$
\\
\hspace{-15pt}
$^1$Niantic
\hspace{20pt}
$^2$UCL\\
{\color{magenta} \url{www.github.com/nianticlabs/footprints}
}
}

\maketitle

\begin{abstract}

Understanding the shape of a scene from a single color image is a formidable computer vision task.
However, most methods aim to predict the geometry of surfaces that are visible to the camera, which is of limited use when planning paths for robots or augmented reality agents.
Such agents can only move when grounded on a \emph{traversable surface}, which we define as the set of classes which humans can also walk over, such as grass, footpaths and pavement.
Models which predict beyond the line of sight often parameterize the scene with voxels or meshes, which can be expensive to use in machine learning frameworks.

We introduce a model to predict the geometry of both visible and occluded traversable surfaces, given a single RGB image as input.
We learn from stereo video sequences, using camera poses, per-frame depth and semantic segmentation to form training data, which is used to supervise an image-to-image network.
We train models from the KITTI driving dataset, the indoor Matterport dataset, and from our own casually captured stereo footage.
We find that a surprisingly low bar for spatial coverage of training scenes is required.
We validate our algorithm against a range of strong baselines, and include an assessment of our predictions for a path-planning task.
\end{abstract}

\section{Introduction}

\begin{figure}[!t]
  \centering
 \includegraphics[width=\columnwidth]{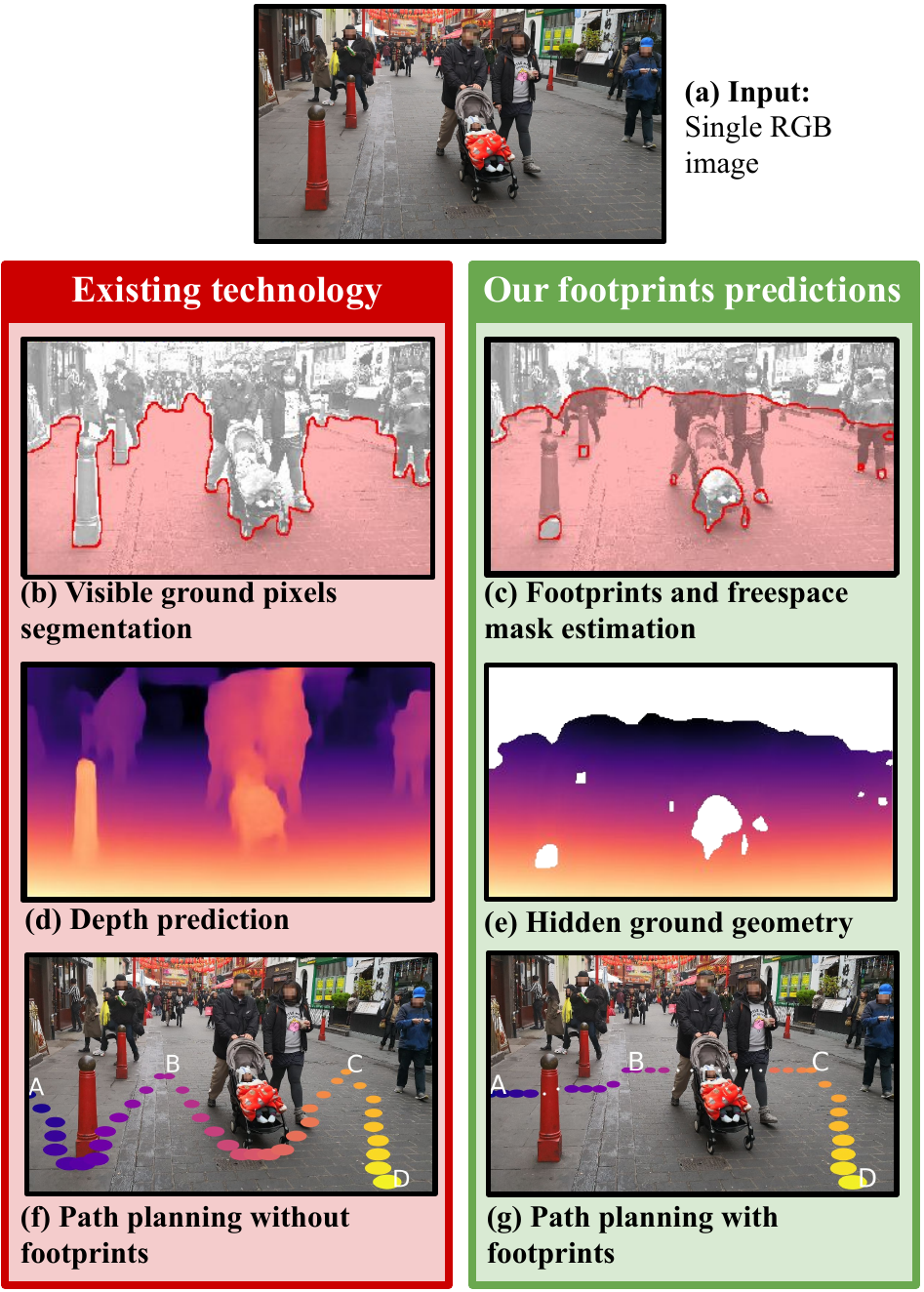}
 \caption{\textbf{Footprints overview:}
    Given a single color image (a), existing methods can estimate the segmentation of which \emph{visible} pixels can be traversed by a human or virtual character (b) and the depth to each pixel (d).
    We introduce \textbf{Footprints}, a method to estimate the extent (c) and geometry (e) that includes \emph{hidden} walkable surfaces.
    Our predictions can be used, for example, to plan paths through the world.
    Here we plan a path from $A \rightarrow B \rightarrow C \rightarrow D$ using the ground predictions, with the A* algorithm \cite{hart1968formal}.
    The baseline path (f) takes an unrealistic route sticking only to \emph{visible ground surface}. Our hidden geometry predictions enable a realistic path behind objects to be found (g).}
    \label{fig:teaser}
    \vspace{-14pt}
\end{figure}

Computerized agents, for example a street cleaning robot or an augmented reality character, need to know how to explore both the \emph{visible} and the \emph{hidden, unseen} world.
For AR agents, all paths must be planned and executed without camera egomotion, so no new areas of the real scene are revealed as the character moves.
This makes typical approaches for path planning in unknown \cite{stein2018learning, wayne2018unsupervised} and dynamic environments  less effective.

We introduce Footprints, a model for estimating both the visible and hidden traversable geometry given just a single color image (Figure \ref{fig:teaser}).
This enables an agent to know where they can walk or roll, beyond the immediately visible surfaces.
Importantly, we model not just the surfaces' overall shapes and geometry  \cite{guo2012beyond, guo2013support}, but also where moving and static objects in the scene preclude walking. We refer to these occupied regions of otherwise traversable surfaces as object \textit{footprints}.

Previous approaches rely on bounding box estimates \cite{hedau2009recovering, lee-nips-2010, schwing2013box}, which are limited to cuboid object predictions.
Other approaches to estimating missing geometry have required \emph{complete}, \emph{static} training environments, which have either been small in scale \cite{FirmanCVPR2016} or synthetic \cite{choy20153dr2n2, song2017semantic}.
Surprisingly, our method can create plausible predictions of hidden surfaces given only partial views of real moving scenes at training time.
We make three contributions:
\begin{enumerate}[topsep=2pt,itemsep=-1ex,partopsep=1ex,parsep=1ex]
    \item We introduce a lightweight representation for hidden geometry estimation from a single color image, with a method to learn this from video depth data.
    \item We present an algorithm to learn from videos with moving objects and incomplete observations of the scene, through masking of moving objects, a prior on missing data, and use of depth to give additional information.
    \item We have produced human-annotated hidden surface labels for all 697 images in the KITTI test set~\cite{Geiger2012CVPR}.
    These are available to download from the project website.
    We also introduce evaluation methods for this task.
\end{enumerate}

\section{Related Work}

Our method is related to prior work in robotics, path planning, and geometry estimation and reconstruction.

\subsection{Occupancy maps and path planning}

If multiple camera views of a scene are available, camera poses can be found and a 3D model of a static scene can be reconstructed \cite{newcombe2010live}. The addition of a segmentation algorithm enables the floor surface geometry to be be found \cite{bao2014understanding, mccormac2017semanticfusion}.
In our work, we make floor geometry predictions given just a single image as input.
Other multi-view approaches include occupancy maps in 2D \cite{senanayake2018building} and 3D \cite{newcombe2011kinectfusion, triebel2006multi, wurm2010octomap}, where new observations are fused into a single map.

The planning of paths of virtual characters or robots in environments with known geometry is a well-studied problem \cite{chong2009robot, gerstweiler2018dargs, kuffner1998goal, santana2011approach, stentz1997optimal}.
Our prediction of walkable surfaces beyond the line of sight shares concepts with works which allow for path planning in environments where \emph{not all geometry can be observed} \cite{stein2018learning, wayne2018unsupervised}.
Gupta \etal \cite{gupta2017cognitive} learn to plan paths with a walkable geometry belief map similar to our world model, while \cite{kumar2019learning} learn potential navigable routes for a robot from watching video.
Rather than directly planning paths, though, in our work we directly learn and predict geometry, which is useful for path planning and more.

\subsection{Predicting geometry you \emph{can} see}

A well-studied task for geometry estimation is the prediction of a depth map given a single color image as input.
The best results here come from supervised learning, \eg \cite{eigen2014depth, fu2018deep}.
Acquiring supervised data for geometry estimation is hard, however, so a popular approach is self-supervised learning, where training data can be monocular \cite{godard2019digging, ranjan2018adversarial, zhou2017unsupervised} or stereo \cite{garg2016unsupervised, godard2017unsupervised, poggi20183net, xie2016deep3d} images.
Depths are learned by minimising a reprojection loss between a target image and a warped source view.
Like these works, we also learn from arbitrary videos to predict geometry, but our geometry predictions extend \emph{beyond the line of sight of the camera}.

\subsection{Predicting geometry you \emph{can't} see}

We fall into the category of works which predict geometry for parts of the scene which are not visible in the input view.
For example, \cite{pan2019cross, song2018im2pano3d} perform view extrapolation, where semantics and geometry outside the camera frustum are predicted.
In contrast, we make predictions for geometry which is \emph{inside} the camera frustum, but which is \emph{occluded} behind objects in the scene.

\paragraph{Geometry completion} Predicting the occupancy of unobserved voxels from a single view is one popular representation for hidden geometry prediction \cite{choy20153dr2n2, FirmanCVPR2016, song2017semantic}.
Training data for dense scene completion is difficult to acquire, though, often making synthetic data necessary \cite{choy20153dr2n2, song2017semantic}. Further, voxels can be slow to process and computation hard to scale for geometry prediction, making their use in real-time or on mobile platforms difficult.
Meshes are a more lightweight representation \cite{silberman2014contour} but incorporating meshes in a learning framework is still an active research topic; a typical approach is to go via an intermediary voxel representation, \eg \cite{meshrcnn}.
A complementary source of information is physical stability as a cue to complete scenes \cite{shao2014imagining}.

\paragraph{Layered completion}
Recent works have taken a lightweight approach to predicting hidden scene structure by decomposing the visible image into layers of color and depth behind the immediately visible scene \cite{dhamo2018peeking, liu2016layered, shin2019multilayer, tulsiani2018layer}.
Similarly, amodal segmentation \cite{follmann2019learning, qi2019amodal, zhu2017semantic} aims to predict overlapping semantic instance masks which extend beyond the line of sight.
However, amodal segmentation doesn't label the contact points necessary to know the location of objects. Amodal segmentation would label a `traversable surface' as continuous under a car or person.

\begin{figure}[!t]
  \centering
 \includegraphics[width=1.0\columnwidth]{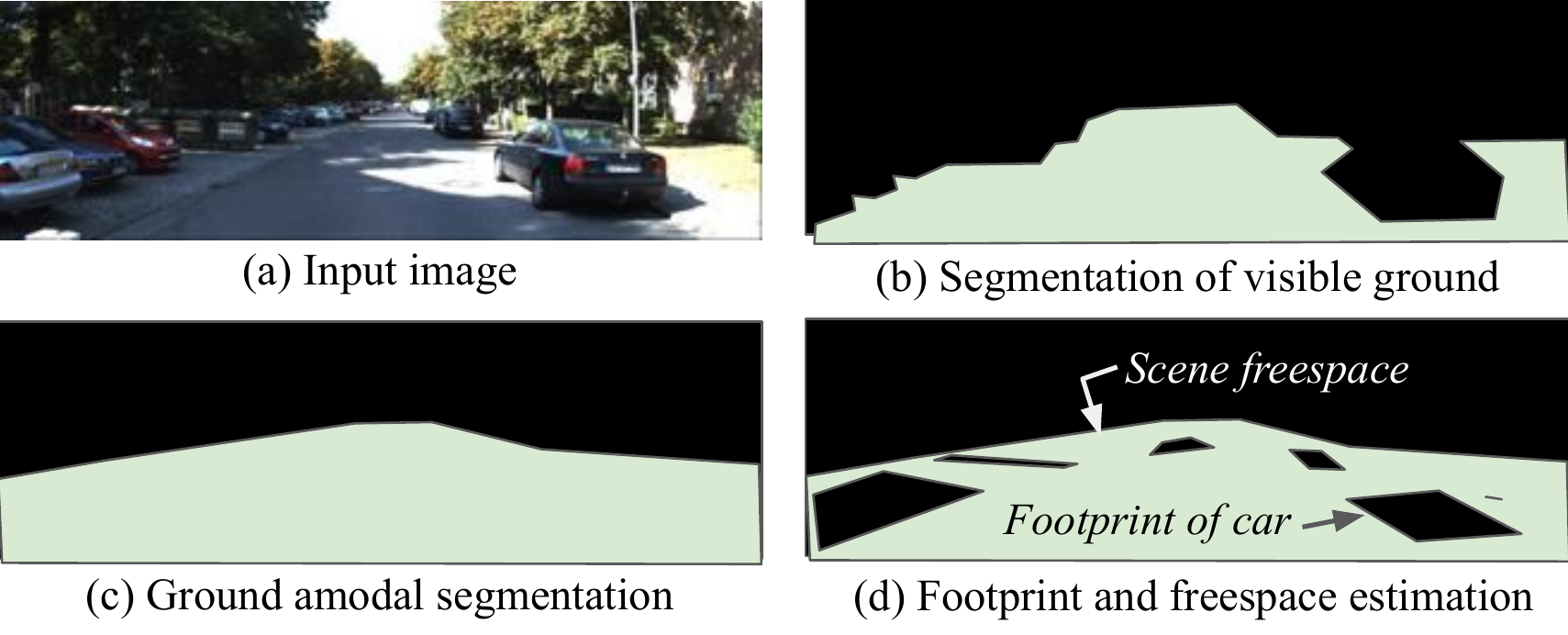}
 \caption{For an input image, segmentation (b) only captures traversable surface visible from this viewpoint, while amodal segmentation (c) fails to delineate which parts of the ground cannot be traversed due to the presence of objects. Our goal (d) is to capture the free, traversable space in the scene and the footprints of objects which preclude motion.}
 \label{fig:different_to_amodal}
 \vspace{-7pt}
\end{figure}

\paragraph{Floor map prediction}
Similar to amodal segmentation are approaches that predict the floor map from a single color image, for example \cite{schulter2018learning, wang2019parametric}.
Similarly \cite{guo2012beyond, guo2013support} complete support surfaces in outdoor and indoor scenes respectively.
The aim of these approaches is to predict support surfaces as if all objects were absent (Figure~\ref{fig:different_to_amodal}(c)), akin to amodal segmentation, while we aim to predict the walkable floor surface taking obstacles into account (Figure~\ref{fig:different_to_amodal}(d)).
The Manhattan layout assumption can be useful to help infer the ground surface in indoor scenes (\eg \cite{hedau2009recovering, lee2017roomnet, lee-nips-2010, schwing2013box}), however, is less applicable outdoors.
Our task is motivated by prior work \cite{watson2018dissertation}, though our approach is novel.

\paragraph{Detection approaches}
One method to estimate the full extent of partially observed objects is via 3D detection, for example 3D bounding boxes \cite{ku2019monocular,  licvpr2019, lin2013holistic, roddick2019orthographic, xiao-cvpr-2016}.
Generic object bounding box detectors have been used to estimate indoor free space \cite{hedau-cvpr-2012, lee-nips-2010, schwing2013box}.
Bounding boxes only give convex footprints for `things' in the image, so aren't suitable for the geometry of `stuff' \cite{caesar2018coco} such as walls, piles of items, or shrubbery.
To the best of our knowledge, object detection has not been effectively combined with amodal segmentation to give traversable surfaces.
We compare to recent object detection baselines and show that our approach is better suited to our task (Section~\ref{sec:experiments}).
Another detection approach is to fit 3D human models to help estimate the hidden layout \cite{fouhey2014people, monszpart2018imapper}, while our aim also has similarities to \cite{gupta-cvpr-2011}, who aim to recover the places in a scene a human can stand, sit and reach.
Such methods often operate with a static scene assumption and work best when the whole scene has been ``explored'' by the humans.

\vspace{4pt}

In comparison to these related works, we predict the hidden and visible traversable surfaces from a single image, taking all obstacles (whether `things' or `stuff') into account.

\section{Our Footprints world model}

Our goal is to predict both the visible and hidden traversable surface for a single color image $I_t$.
A surface is defined as traversable if it is visually identifiable as one of a predefined set of semantic classes, listed in our  supplementary material.
The \textbf{visible traversable surface} can be represented with two single-channel maps: \vspace{-4pt}
\begin{enumerate}\setlength\itemsep{-0.4em}
    \item A visible ground segmentation mask $S$. Each $s_{j} \in S$ is $1$ if the surface seen at pixel $j$ is from a traversable class, and $0$ otherwise. $S$ can be estimated with \eg \cite{harakeh2016identifying, yao2015estimating}.
    \item A visible depth map $D$ giving the distance from the camera to each visible pixel in the scene, \eg \cite{godard2017unsupervised}.
\end{enumerate}
Together, $\{D, S\}$ model the extent and geometry of all the \emph{visible} ground which can be traversed -- Figure \ref{fig:different_to_amodal}(b).
However, to know about how an agent could move through areas of the scene beyond the line of sight, we also need to model geometric information about ground surfaces which are occluded by objects.
To this end, our representation also incorporates two channels which model the \textbf{hidden traversable surface}: \vspace{-3pt}
\begin{enumerate}\setlength\itemsep{-0.4em}
  \setcounter{enumi}{2}
    \item A \emph{hidden} ground segmentation mask $S^*$, which represents the extent of the entire traversable floor surface inside the camera frustum, including occluded parts.
    Each pixel $s^*_{j} \in S^*$ is $1$ if the camera ray associated with pixel $j$ intersects with a \emph{walkable} surface at any point (even behind objects visible in this view) and $0$ otherwise.
    This can also be seen as a top-down floor map reprojected into the camera view \cite{gupta2017cognitive}.
    \item A depth map $D^*$ which gives the geometry of the hidden ground surface.
    Each $d^*_{j} \in D^*$ contains the depth from the camera to the (visible or hidden) ground for pixel $j$.
    If the camera ray at pixel $j$ doesn't intersect any traversable surface (\ie $s^*_{j}=0$), then $d^*_{j}$ is $0$.
\end{enumerate}

Our four-channel representation $\{S, D, S^*, D^*\}$ is a rich world model which enables many tasks in robotics and augmented reality, while being lightweight and able to be predicted by our standard image-to-image network.

\begin{figure*}
  \begin{flushright}
 \includegraphics[width=1.0\textwidth,trim={0.9cm 2cm 2.5cm 1cm},clip]{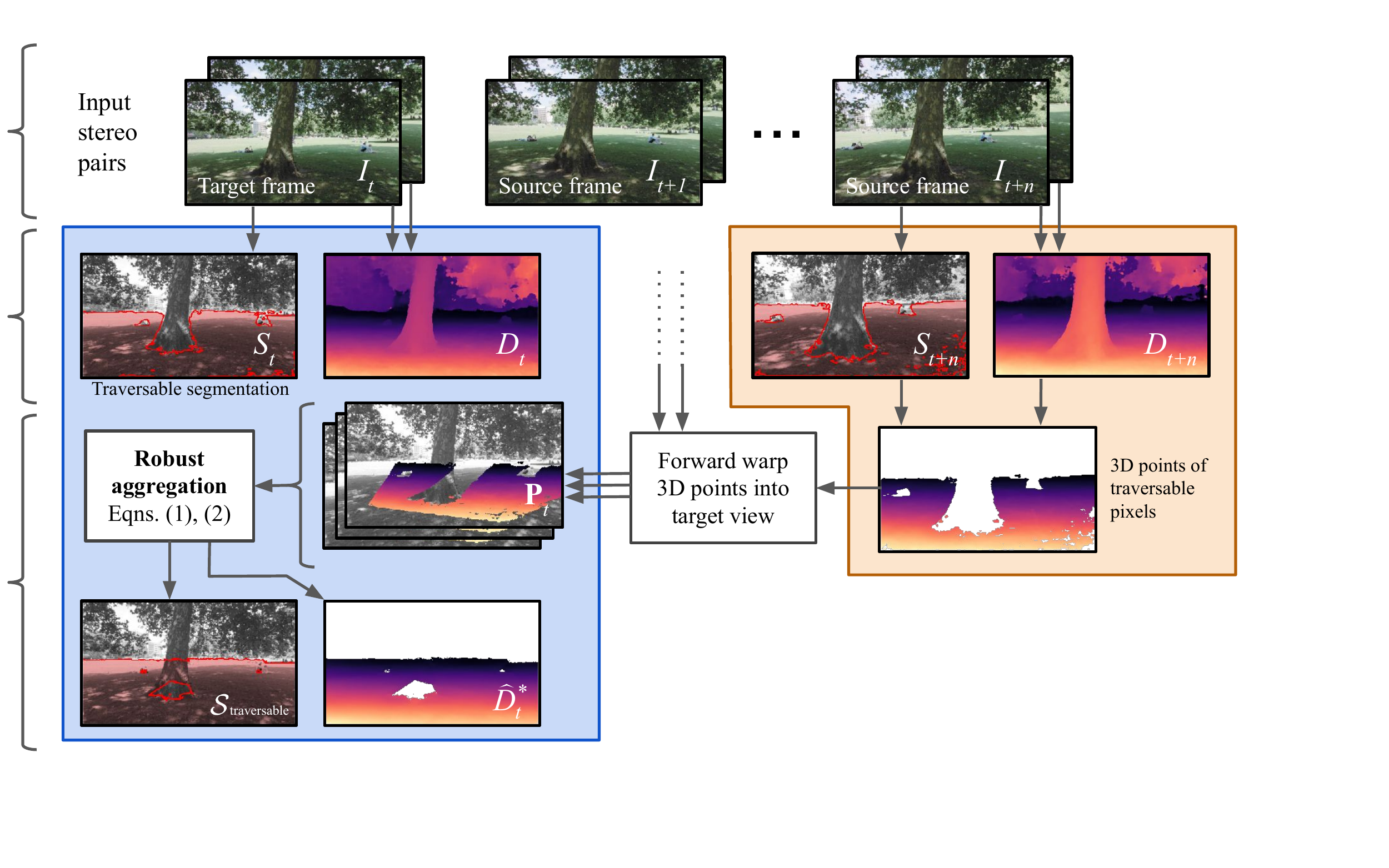}
  \begin{minipage}[r]{0.52\textwidth}
  \vspace{-90pt}
  \caption{\textbf{Generating training data from multiple views:} For a frame $I_t$, the visible traversable surface map $S_t$ is estimated directly from the image, while depth $D_t$ is estimated from the stereo pair. Hidden ground surface information about the target frame is generated from source frames $I_{t+n}$.
  Segmentation masks from these source frames are projected back into the target frame, and used to generate $\mathcal{S}_\text{traversable}$ and $\hat{D}^*_t$.}
  \end{minipage}
  \end{flushright}
 \label{fig:how_to_train}
 \vspace{-30pt}
\end{figure*}

\begin{description}\setlength\itemsep{-0.2em}

    \item[How does our model relate to ground segmentation?]
    A semantic segmentation algorithm also gives us the pixels which an agent could walk on, but only those which are visible by the camera (\ie $S$).
    Our model also represents the location of walkable ground surfaces which are not visible to the camera.

    \item[Why can't we just fit a plane?]
    Assuming a planar floor surface, fitting a plane to the visible ground would give an estimate of the \emph{geometry} of the walkable surface.
    However, this planar model does not give the \emph{extents} of the walkable surface, meaning an agent traversing the scene would walk into objects.

    \item[Why not use a voxel model?]
    Our image-space predictions are lightweight and memory efficient, and furthermore output is pixel-wise aligned with the input space.
    Given that our main focus is on where we can walk, our representation is the minimal necessary representation.

    \item[Why not make the predictions in top-down space?]
    We could represent the world in top-down view instead of in reprojected camera space.
    While this would allow us to model the world outside the camera frustum, we would add complexity, with more complicated training and reliance on good test-time camera-pose estimation.



\end{description}

\section{Learning to predict Footprints}
It is possible to estimate $\{S, D\}$ using off-the-shelf prediction models, \eg \cite{kendall2018multi}.
However, training a model to estimate $\{S^*, D^*\}$ requires additional sources of information.
Human labeling is expensive and difficult to do at scale as we are asking an annotator to label occluded parts of a scene.
Instead, we exploit two readily available sources of information: freely captured video and depth data.
We use these to divide pixels from each training image into three disjoint sets.
$\mathcal{S}_\text{traversable}$ contains indices of pixels which are deemed to be traversable;
$\mathcal{S}_\text{untraversable}$ the indices of pixels which we are confident cannot be traversed,
and $\mathcal{S}_\text{unknown}$ the indices of pixels which we have no information about.
These \emph{unknown} predictions come about by our use of freely captured video for training; some areas of the scene have never been observed, and we have no information about whether these regions are traversable or not.

\subsection{Learning $\mathcal{S}_\text{traversable}$ from video data}
Freely captured video is easy to obtain and gives us the ability to generate training data for geometry behind visible objects.
We use other frames in the video to provide information about what the geometry and shape of the walkable surface is by projecting observations from each frame back into the target camera.

We use off-the-shelf tools to estimate camera intrinsics and depth maps for each frame and relative camera poses between source frames $I_{t+i}$ and the target frame $I_t$.
We then forward warp \cite{schwarz2007non, wang2018occlusion} the depth values of traversable pixels from the source frame into the target frame. 
This results in a sparse depth map $P_{t+i\xrightarrow{} t}$, representing the geometry and extents of the traversable ground visible in frame $I_{t+i}$ rendered from the viewpoint of $I_t$.
We repeat this forward-warping for $N$ nearby frames,  obtaining the set $\textbf{P}_t = \{P_{t+i\xrightarrow{} t}\}_{i=1}^N $.

Due to inaccuracies in floor segmentation, depth maps, and camera poses, many of the reprojected floor map images $P_{t+i\xrightarrow{} t}$ will have errors.
We therefore perform a \mbox{\emph{robust}} aggregation of the multiple noisy segmentation and depth maps to form a single training image.
Our traversable labelset $\mathcal{S}_\text{traversable}$ is formed from pixels for which at least $k$ reprojected depth maps contain a nonzero value, \ie
\begin{align}
    \mathcal{S}_\text{traversable} &= \bigg\{ j \in \mathcal{J} \; | \;
    \Big( \sum_{P \in \textbf{P}_{t}}    \left[ p_j > 0 \right] \Big) > k \bigg\},
\end{align}
where $[]$ is the Iverson bracket, $\mathcal{J}$ is the set of all pixel indices in this image and $p_j$ is the $j$th pixel in $P$.
See \mbox{Figure~\ref{fig:how_to_train}} for an overview.

We subsequently obtain our ground depth map $\hat{D}^*$ by taking the median depth value (ignoring zeros) associated with each pixel $j$ if and only if there is a valid depth value at this location: 
\begin{align}
    \hat{D}^* &= \text{median} \big( \big\{ P \in \textbf{P}_{t} \; | \; P > 0 \big\} \big)
\end{align}
We supervise our prediction $\hat{D}^*$ with a $\log L_1$ loss \cite{Hu2018Revisiting}.

\subsection{Depth masking to find $\mathcal{S}_\text{untraversable}$}

While $\textbf{P}_t$ is constructed from depth images of multiple source images, models trained on $\textbf{P}_t$ alone typically incorrectly estimate object footprint boundaries, often entirely missing the footprints of thin objects such as poles and pedestrians. Such mistakes are due to inaccuracies in camera pose tracking, traversable segmentation masks and visible depth maps, resulting in sometimes poor reprojections into the target frame that are not excluded by our robust aggregation method.
To tackle this problem, we exploit depth data from the target image $I_t$ to estimate $\mathcal{S}_\text{untraversable}$, the set of pixels in the image which are \emph{definitely not} traversable. Subsequently, we redefine $\mathcal{S}_\text{traversable}$ to not include pixels in $\mathcal{S}_\text{untraversable}$.

\begin{figure}[!t]
  \centering
 \includegraphics[width=\columnwidth]{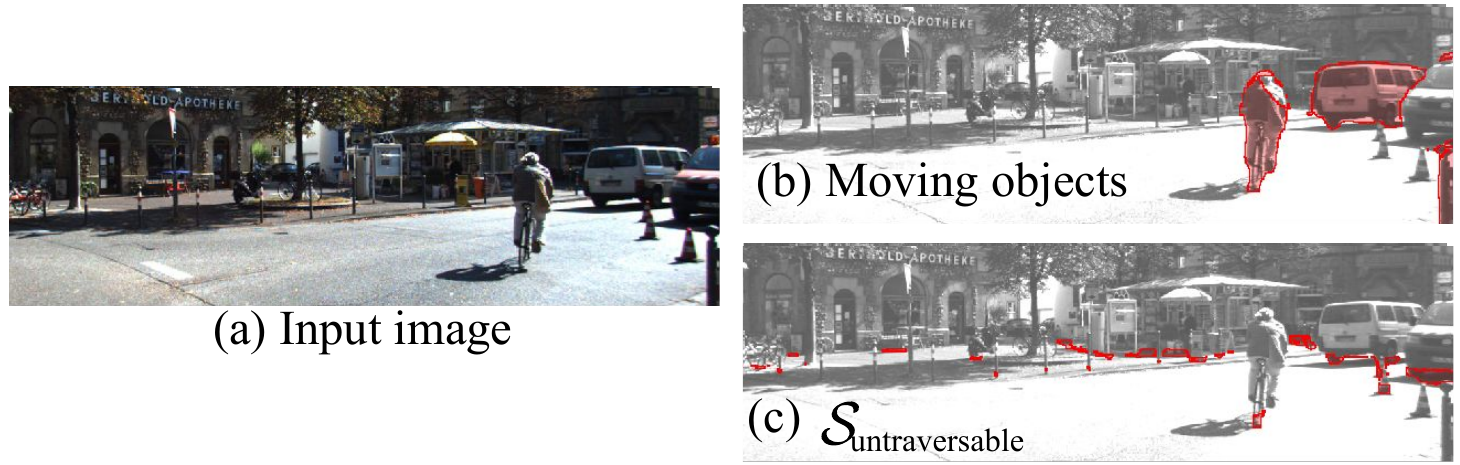}
 \caption{\textbf{Moving object and depth masking:} For a training image (a), our moving object mask (b) identifies pixels associated with moving objects. The set of pixels $\mathcal{S}_\text{untraversable}$ (c) uses the training depth image to capture the footprints of small and thin objects.
 \label{fig:masks}}
 \vspace{-5pt}
\end{figure}

To find $\mathcal{S}_\text{untraversable}$, we first project all points in the depth map $D_t$ from camera space into world space. Next, we fit a plane to those points which are classified as visible ground in our segmentation mask $S_t$ using RANSAC \cite{ransac1981}. We then move each point in the world along the normal vector of the plane such that they now lie on the plane, and `splat' in a small grid around the resulting position. After reprojecting these points back into camera space, we apply a filtering step (see supplementary material for details) to remove erroneous regions, and obtain the set of pixels $\mathcal{S}_\text{untraversable}$.
An example is shown in Figure~\ref{fig:masks}(c).

\subsection{Masking moving objects at training time}

The computation of $\mathcal{S}_\text{traversable}$ and $\hat{D}^*$ utilizes multiple frames and makes the significant and unrealistic assumption that our training data comes from a static world, when in fact many objects will undergo significant motion between frames $I_t$ and $I_{t+i}$.
To combat this, we identify and remove pixels from our training loss which are associated with moving objects.
We could use semantic segmentation to remove non-static object classes, such as cars; however, this would prevent us learning about the hidden geometry of \emph{any} cars, including parked ones.
We could train on static scenes \cite{li2019mannequin}, but would be limited by the availability of existing general-purpose datasets.
We instead observe that most classes of moving objects are static at least some of the time.
For example, while it is hard to learn the geometry of a moving vehicle, we can learn the shape of parked cars and apply this knowledge to moving cars at test time.
Similarly, footprints of humans can be learned by observing those which are relatively static in training.

We compute a per-pixel binary mask $M$, where $\mu_j \in M$ is zero for pixels depicting non-static objects.
To compute $M$ for frame $t$, we computed the \emph{induced flow} \cite{wang2019unos, zou2018df} from frame $t$ to $t+1$, using $D_t$ and camera motion.
This estimated where pixels would have moved to assuming a static scene.
We also separately estimate frame-to-frame optical flow.
Pixels where the induced and optical flow differ are often pixels on moving objects; we set $\mu_j$ to $0$ if the endpoints of the two flow maps differ by more than $\tau = 3$ pixels, and $1$ otherwise.
An example of $M$ is shown in Figure~\ref{fig:masks}(b).

\subsection{Final training loss}
\label{sec:final_loss}
Our training loss comprises four parts, one for each output channel $\{S^*, D^*, S, D\}$.

\begin{description}
    \item[Hidden traversable surface loss $l^{s^*}_j$] --- \vspace{-3pt}
\begin{subnumcases}{l^{s^*}_j=}
    -\mu_j \log( \hat{s^*_j} )
        & \hspace*{-1em} if $j\in \mathcal{S}_\text{traversable} $ \label{eqn:traversable} \\
    -\log(1 - \hat{s}^*_j)
        &  \hspace*{-1em} if $j \in \mathcal{S}_\text{untraversable}$ \label{eqn:untraversable}\\
    - \lambda \log( 1 - \hat{s^*_j} )
        & \hspace*{-1em} otherwise, \label{eqn:unknown}
\end{subnumcases}
where (\ref{eqn:traversable}) encourages pixels in $\mathcal{S}_\text{traversable}$ to be labelled $s^*_j = 1$;
(\ref{eqn:untraversable}) encourages pixels in $\mathcal{S}_\text{untraversable}$ to be labelled $s^*_j = 0$;
and (\ref{eqn:unknown}) applies a prior $\lambda < 1$ to the remaining, unknown pixels which conservatively encourages them to be labeled as untraversable.

\item[Visible traversable surface loss $l_j^s$] ---
This is supervised using standard binary cross-entropy loss.

\item[Observed depth loss $l_j^d$] ---
For the channel predicting the depth map for visible pixels, we follow \cite{Hu2018Revisiting, watson2019depthints} and supervise with $l_j^d = \log(|d_j - \hat{d}_j| + 1)$.
\item[Hidden depth loss $l_j^{d^*}$] ---
Hidden depths are also supervised with the log $L_1$ loss, but we only apply the loss for pixels $\in \mathcal{S}_\text{traversable}$.
\end{description}
Our final loss is the sum of each subloss over all pixels:
\begin{align}
    L &= \sum_j l_j^{s^*} + l_j^s + l_j^d + l_j^{d^*}.
\end{align}

\subsection{Implementation details}

To generate training signals for KITTI and our casually captured stereo data, camera extrinsics and intrinsics are estimated using \mbox{ORB-SLAM2}~\cite{mur2017orb2}, while depth maps are inferred from stereo pairs using \cite{PSMNet}. Segmentation masks are estimated using a simple image-to-image network trained using the ADE20K \cite{ADE20K} and Cityscapes \cite{CityscapesDataset} datasets, and optical flow is estimated using \cite{hui18liteflownet, pytorch-liteflownet}. 
Our network architecture is based on \cite{godard2019digging}, modified to predict four sigmoided output channels.
We adjust our training resolution to approximately match the aspect ratio of the training images: $512\times 640$ for the Matterport dataset, $192\times640$ for KITTI, and $256\times448$ for our own stereo data.
For Matterport, camera intrinsics, relative locations, and depth maps are provided. Thus we need only estimate segmentation masks, and do so using the same pretrained network finetuned on a small subset of \mbox{5,000} labelled Matterport images.
Except in some ablations, we set $\lambda = 0.25$.

\begin{figure}
  \centering
 \includegraphics[width=\columnwidth]{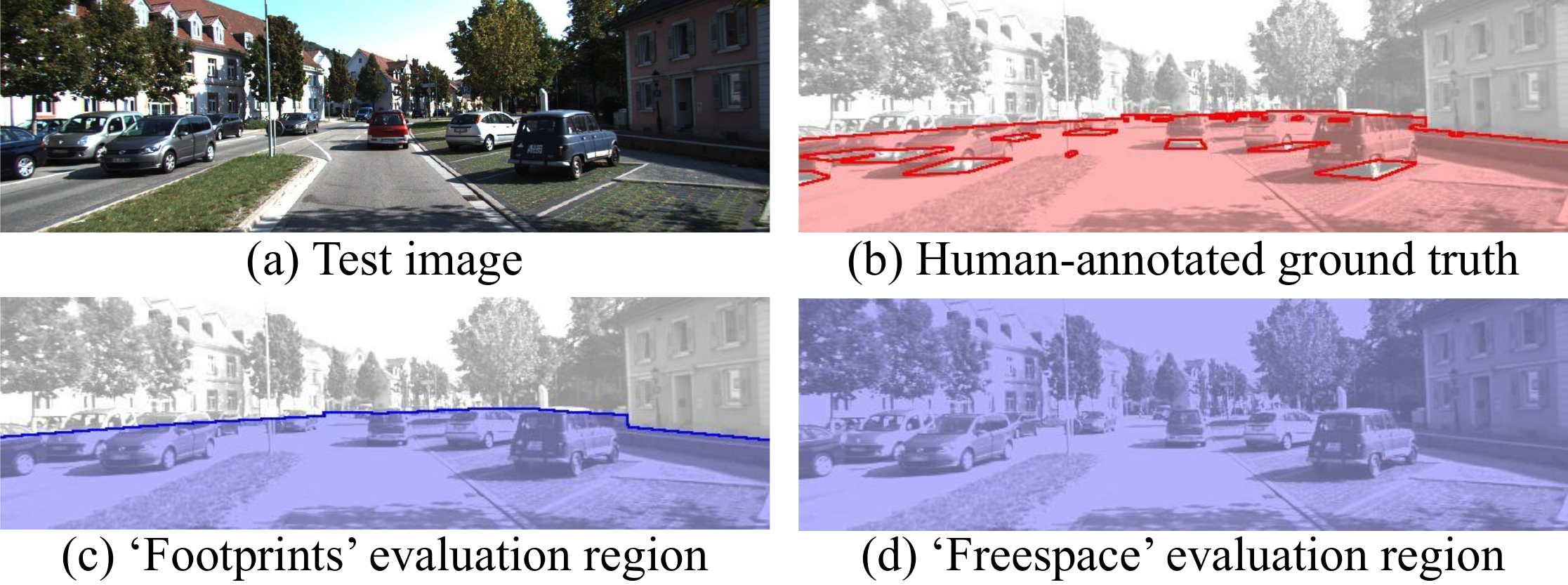}
 \caption{\textbf{Evaluation region:} Object footprints are evaluated on all pixels within the human-annotated ground polygon (a), while for freespace evaluation we use the whole image (b); see \mbox{Sec~\ref{sec:experiments}}.
 \label{fig:eval_region}}
 \vspace{-5pt}
\end{figure}

\begin{figure*}[!t]
  \centering
    \resizebox{1.0\textwidth}{!}{
        \input{figs/kitti_results/index.tex}
    }
 \caption{\textbf{KITTI results:} Each row shows an input image and the predicted $S^*$ mask from our model, our model without depth masking and the strongest baseline.
 We find the footprints of a wider variety of objects than the baseline, and are better at capturing the overall shape of the traversable space.
 The 1st and 4th rows show the benefits of depth masking for thin objects.
 The final row shows a failure, where we fail to predict walkable space behind a car; our facing-forward training videos means our network rarely sees behind some objects.}
 \label{fig:qualitative_results}
 \vspace{-12pt}
\end{figure*}
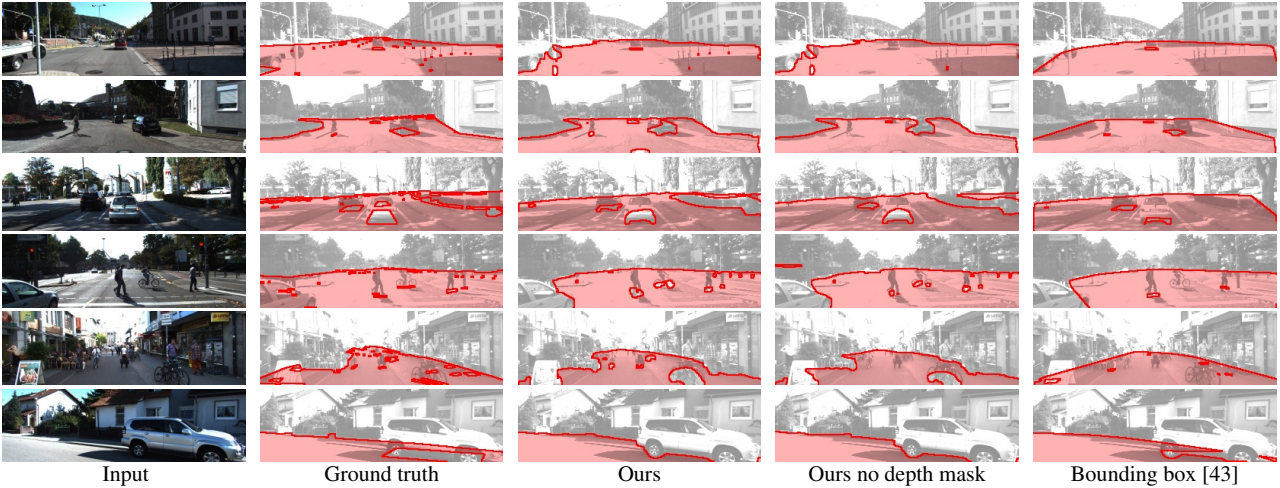

\section{Experiments}
\label{sec:experiments}

We validate our scene representation and the associated learning approach experimentally. We do this by:
\begin{enumerate}[label=\alph*)]\vspace{-3pt}
  \setlength\itemsep{-0.3em}
  \item Quantifying the accuracy of our predictions indoors and outdoors (\textbf{Matterport} and \textbf{KITTI}),
  \item Illustrating their quality across different scenarios,
  \item Ablating to gauge the benefits of different design decisions, and
  \item Illustrating a use-case where Footprints are used for path planning (Sec~\ref{Sec:PathPlanning}).
\end{enumerate}
We focus our evaluation here on hidden traversable surface estimation \ie, $S^*$, and we evaluate $S$, $D$ and $D^*$ in the supplementary material.

\paragraph{Metrics:}
There are two aspects of  $S^*$  predictions which are of interest:
(1) The ability to estimate the overall extents of the traversable freespace in the image, and (2) the ability to estimate the footprint base of objects in the scene which must be avoided.
To capture this we introduce two evaluation settings.
The first, \emph{freespace evaluation}, addresses (1) by evaluating our thresholded prediction of $S^*$ over all pixels in the image using the standard binary detection metrics of IoU and F1.
The second is \emph{footprints evaluation} addressing (2), where we focus on the evaluation of object footprints by evaluating only within the ground region.
To evaluate all methods equivalently, we evaluate within the true ground segmentation (KITTI) and the convex hull of the true visible ground (Matterport) --- see Figure~\ref{fig:eval_region}.

\paragraph{Baselines:}
We compare against several baselines, to demonstrate the efficacy of our method across tasks:

\paragraph{Visible only} ---  $S^*$ is set as the visible ground mask $S$.

\paragraph{Convex hull}  --- We estimate $S^*$ as the convex hull of the visible ground mask  $S$.

\paragraph{3D bounding boxes} ---  Footprints of objects are estimated  using 3D bounding box detectors \cite{mousavian20173d} for outdoor scenes and \cite{qi2019deep} indoors; we evaluate both the `ScanNet' and `SunRGBD' models from \cite{qi2019deep}.
Estimated object footprints are subtracted from the convex hull baseline for the final prediction.
Unlike our method, \cite{qi2019deep} make predictions with access to the structured-light-inferred depth map at test time; we include their state-of-the-art results as an upper bound on what a bounding box method could achieve.

\paragraph{Voxel prediction} --- On indoor scenes, we use \cite{song2017semantic} to estimate the voxelized scene from a depth input.
Voxels estimated as `floor' are reprojected into the camera.

\paragraph{Project down} --- We train a model to estimate footprints using only depth images at training time, without our multi-frame reprojection.
For this we train a binary classifier to predict if it expects each pixel to be a member of $\mathcal{S}_\text{untraversable}$ or not, and subtract these pixels from the convex hull.

\begin{table}
  \centering
  \small
  \resizebox{1.0\columnwidth}{!}{
\begin{tabular}{lcccc}
    & \multicolumn{2}{c}{Freespace eval.} & \multicolumn{2}{c}{Footprint eval.} \\
    \cmidrule(lr){2-3}\cmidrule(lr){4-5}
    & IoU & F1 & IoU & F1    \\
    \shline

Convex hull                                 &  0.790   &  0.876    &  0.145  & 0.230 \\
Bounding box \cite{mousavian20173d}          &  \underline{0.794}   &  \underline{0.879}    &  0.187  & 0.292 \\
Nothing traversable  ($S^* = 0$)            &  0.000   &  0.000    &  0.089  & 0.153 \\
Everything traversable  ($S^* = 1$)         &  0.344   &  0.506    &  0.000  & 0.000 \\
Visible ground                              &  0.770   &  0.860    &  \underline{0.231}  & \underline{0.356} \\
\textbf{Ours}                               &  \textbf{0.797}   &  \textbf{0.880}    &  \textbf{0.239}  & \textbf{0.363} \\
  \end{tabular}
  }
  \vspace{1pt}
  \small
  \caption{\textbf{Evaluating object footprint and freespace detection on the KITTI dataset:}
  Best methods in each category are \textbf{bolded}; second best \underline{underlined}.
  Our method outperforms all baselines. \label{tab:kitti_compare_to_baselines}}
  \vspace{-2pt}
\end{table}

\begin{table}
  \centering
  \small
  \resizebox{1.0\columnwidth}{!}{
\begin{tabular}{lcccc}
    & \multicolumn{2}{c}{Freespace eval.} & \multicolumn{2}{c}{Footprint eval.} \\
    \cmidrule(lr){2-3}\cmidrule(lr){4-5}
    & IoU & F1 & IoU & F1    \\
    \shline
Project down baseline               &  0.344   &  0.506    &  0.082  & 0.144 \\
Ours w/o moving object masks        &  0.795   &  0.878    &  0.227  & 0.347 \\
\textit{as above} w/o eqn.~(3b)     &  \textbf{0.797}   &  \underline{0.879}    &  0.218  & 0.333 \\
Ours w/o eqn.~(3b)                  &  0.793   &  0.877    &  0.225  & 0.343 \\
Ours ($\lambda = 0$)                & 0.355   &  0.519    &  0.217  & 0.335 \\
Ours ($\lambda = 0.5$)                &  0.787   &  0.873    &  0.232  & 0.355 \\
Ours ($\lambda = 1.0$)                &  0.776   &  0.865    &  \underline{0.234}  & \underline{0.356} \\
\textbf{Ours}                       &  \textbf{0.797}   &  \textbf{0.880}    &  \textbf{0.239}  & \textbf{0.363} \\
  \end{tabular}}
  \vspace{1pt}
  \small
  \caption{\textbf{Ablating our method on the KITTI dataset:} Our ablations validate our approach; removing components of our method gives equivalent \emph{freespace} scores, but are significantly worse at detecting object \emph{footprints}. \label{tab:kitti_ablation}}
  \vspace{-3pt}
\end{table}

\begin{table}
  \centering
  \small
  \vspace{1pt}
  \resizebox{1.0\columnwidth}{!}{
  \begin{tabular}{lccccccccc}
& \multicolumn{2}{c}{Freespace eval.} & \multicolumn{2}{c}{Footprint eval.} \\
    \cmidrule(lr){2-3}\cmidrule(lr){4-5}
    & IoU & F1 & IoU & F1    \\
  \shline
Nothing traversable  ($S^* = 0$)                             &  0.000   &  0.000    &  0.186  & 0.291 \\
Everything traversable ($S^* = 1$)                            &  0.480   &  0.611    &  0.000  & 0.000 \\
\hline
Convex hull                                                  &  0.454   &  0.562    &  0.289  & 0.421 \\
Bounding box$\dagger$ \cite{qi2019deep} (Scannet)            &  0.450   &  0.557    &  0.333  & 0.469 \\
Bounding box$\dagger$ \cite{qi2019deep} (Sun RGBD)           &  0.451   &  0.559    &  0.315  & 0.450 \\
Voxel SSCNet$\dagger$ \cite{song2017semantic}                &  0.492   &  0.615    &  0.087  & 0.136 \\
Voxel SSCNet+$\dagger$                                       &  0.418   &  0.547    &  0.107  & 0.173 \\
Visible ground ($S^* = S$)                                   &  \underline{0.505}   &  \underline{0.628}    &  \underline{0.404}  & \underline{0.542} \\
\textbf{Ours}                                                         &  \textbf{0.652}   &  \textbf{0.767}    &  \textbf{0.426}  & \textbf{0.557} \\
\hline
\textbf{Ours} ($\lambda = 0.5$)                                       &  \textbf{0.663}   &  \textbf{0.776}    &  \textbf{0.452}  & \textbf{0.585} \\
  \end{tabular}
  }
  \vspace{1pt}
  \caption{\textbf{Evaluation of $S^*$ on Matterport \cite{Matterport3D}:}
  We are the best method in both freespace and footprint evaluation across all metrics.
The final row shows an ablation which outperforms \emph{ours}, suggesting that a careful choice of hyperparameters could further improve performance.
  Methods marked $\dagger$ have access to structured-light depth data at test time.
  Voxel SSCNet(+)'s geometry estimation failed on 178 scenes; we ignore these when averaging.
  \label{tab:matterport_results}}
  \vspace{-2pt}
\end{table}

\begin{figure}
    \centering
    \vspace{2pt}
    \includegraphics[width=0.9\columnwidth]{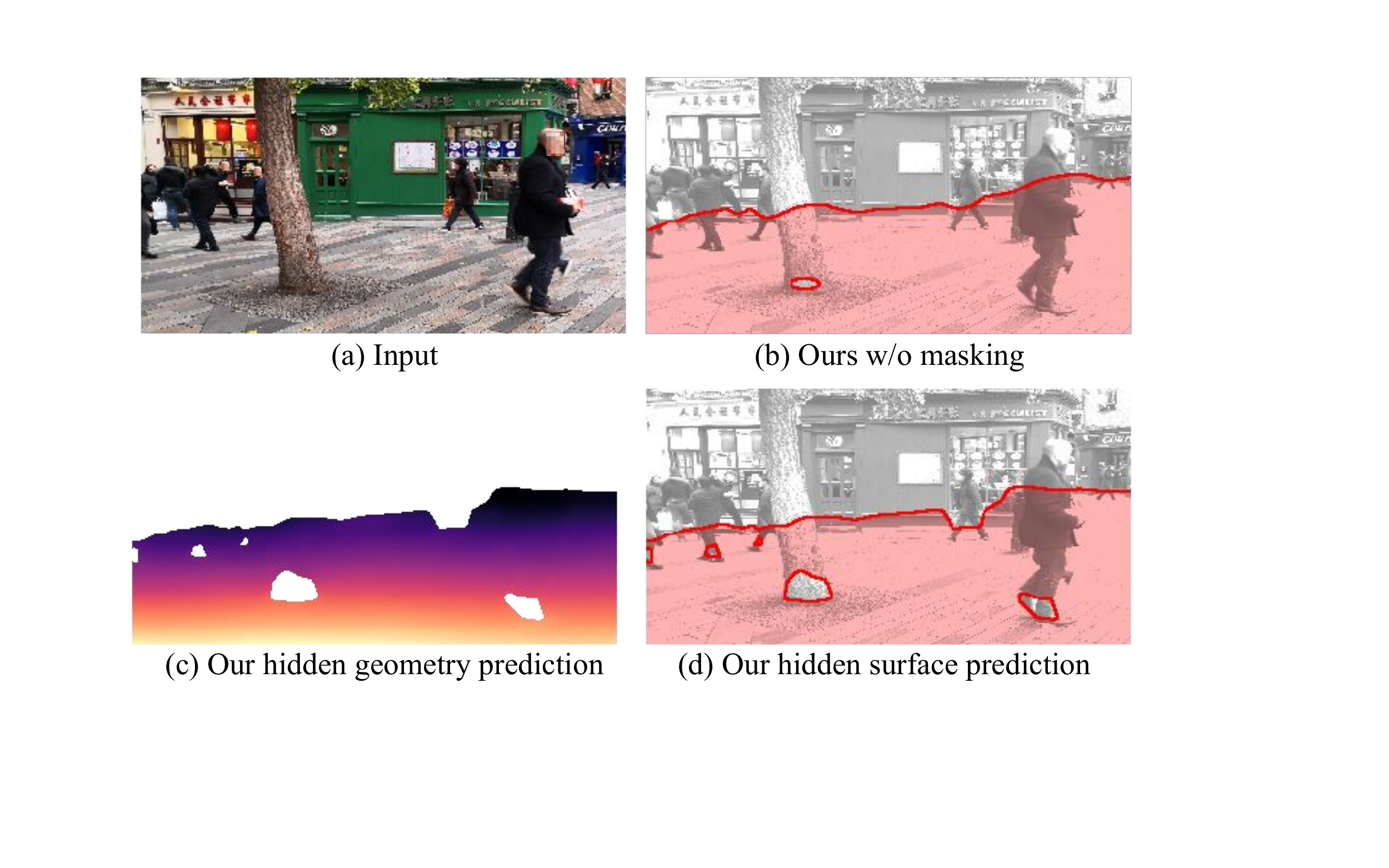}
    \caption{\textbf{Stereo capture results:} Prediction of our model trained on handheld stereo footage for an image taken using a mobile phone. Here we see the qualitative impact of using our full method (c), (d) vs.~with neither depth masking nor moving object masking (b). We are better able to capture the footprints of pedestrians.}
    \label{fig:rig_results}
    \vspace{-26pt}
\end{figure}

\subsection{Evaluation on the KITTI benchmark} \label{ss:kitti_eval}

We first train and evaluate on the well-established \textbf{KITTI} benchmark \cite{Geiger2012CVPR} using the Eigen split \cite{eigen2014depth}.
To evaluate quantitatively, we generate human annotations for the entire test set.
Labelers were instructed to draw a polygon bounding the hidden and visible walkable surface, and to separately label the footprint of each occluding object in the scene.
Due to the nature of our task, labelers had to estimate the hidden extents of many objects, which seems like an error-prone task.
However, this follows work in amodal labeling where consistency between labelings was found to be reasonably high~\cite{qi2019amodal}.
These annotations are available from the project website.

 We present quantitative results of our method alongside baselines in Table~\ref{tab:kitti_compare_to_baselines}. Here we demonstrate the superior performance of our method in both freespace and footprint evaluation. Qualitative results can be seen in Figure~\ref{fig:qualitative_results}.
 We see that \textit{Ours} finds the footprints of a wider variety of objects than \textit{Bounding Box} as we are not limited to predefined classes.
 We also better capture the overall shape of the traversable ground.
 Additionally, we ablate our method in Table~\ref{tab:kitti_ablation}, showing that our full method helps to improve results.

\subsection{Indoor evaluation}

\begin{figure}[!t]
  \centering
    \resizebox{1.0\columnwidth}{!}{
        \input{figs/matterport_results/index.tex}
    }
    \vspace{-12pt}
 \caption{\textbf{Matterport results:} We predict the geometry of objects which do not fall into categories detectable by off-the-shelf object detectors, \eg the pillar in the leftmost column. The rightmost column demonstrates how we can predict the continuation of traversable surfaces through doorways. No SSCNet+ results were computed for the third column due to layout estimation failure.}
 \label{fig:qualitative_results_matterport}
\end{figure}
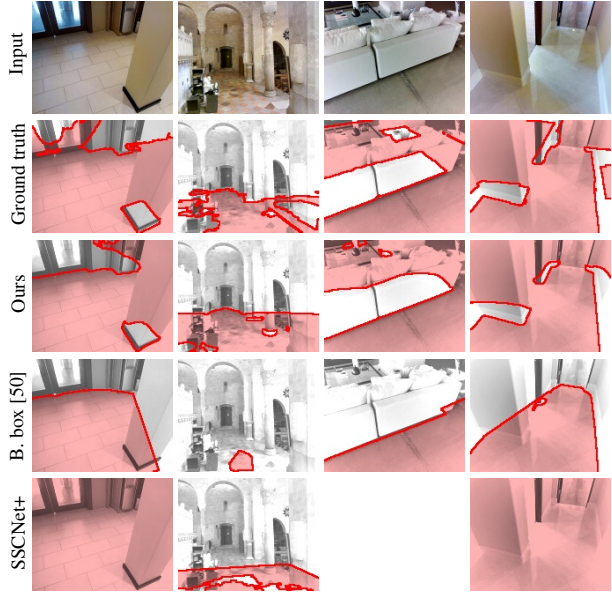

We use the \textbf{Matterport} dataset \cite{Matterport3D} for training and evaluation on indoor scenes.
Here, camera poses and structured-light depth maps are provided, and the ground truth floor masks and geometry are rendered from the dataset's semantically annotated mesh representation.
We only train and evaluate on images from the forward- and downward-facing cameras on their rig, leaving us with 49,286 training images.
We evaluate on the first 500 images from the test set.
Results are shown in Figure~\ref{fig:qualitative_results_matterport} and Tables~\ref{tab:matterport_results} and \ref{tab:matterport_depth_results}, where we again outperform all baselines.
SSCNet \cite{song2017semantic} performs poorly as this method was mainly trained on synthetic data, where the footprints of objects are not separately delineated from the ground plane.
We therefore create a reworking of their method, SSCNet+.
Here, the voxel predictions of chairs, beds, sofas, tables and TVs are projected to the floor and subtracted to give more accurate footprint estimates.
SSCNet+ achieves higher \emph{footprint} scores than SSCNet, but lower \emph{freespace} scores.

\begin{table}[!t]
  \centering
  \footnotesize
  \resizebox{0.85\columnwidth}{!}{\begin{tabular}{ccccc}

                                    & \textbf{$\boldsymbol{a}_{\boldsymbol{1}}$}     & \textbf{RMSE}    & \textbf{Abs.~rel.}    & \textbf{Sq.~rel.} \\
                                    \shline
        SSCNet$\dagger$ \cite{song2017semantic}                &   0.069   &  6.689   &  1.434  & 14.667 \\
        RANSAC plane                                                 &   0.359   &  1.713   &  0.307  & 0.865 \\
        \textbf{Ours}                                                         &  \textbf{0.577}   &  \textbf{1.101}   &  \textbf{0.206}  & \textbf{0.292} \\
        \hline
        RANSAC (oracle*)                                        &    0.351   &  1.693   &  0.306  & 0.821 \\

    \end{tabular}}
    \vspace{2pt}
    \caption{\textbf{Matterport hidden depth ($D^*$) evaluation:} Our method outperforms the baselines, even the artificially boosted method in the final row which has access to ground truth visible ground segmentation and ground truth depths.\label{tab:matterport_depth_results} }
\end{table}

\subsection{Training from handheld camera footage}

We additionally captured a 98,002-frame video dataset from an urban environment with a stereo camera.
A model trained on this dataset allows us to make plausible predictions on images captured from a mobile phone camera on a different day (Figures~\ref{fig:teaser} and \ref{fig:rig_results}).

\subsection{Inference speed}

Table \ref{tab:matterport_inference_speed} compares our inference speed with competing methods.
For a fair comparison, all methods were assessed with a batch size of one.
Our simple image-to-image architecture is significantly quicker than alternatives, lending itself more readily to mobile deployment.

\begin{table}
  \centering
  {
  \resizebox{0.85\columnwidth}{!}{
  \begin{tabular}{lcc}
  & \textbf{Preprocessing (s)} & \textbf{Inference (s)} \\
  \shline

Voxels (SSCNet) \cite{song2017semantic} & 43 & 66 \\

Bounding box \cite{qi2019deep} &  -  & 0.417  \\

Bounding box \cite{mousavian20173d} &  -  & 0.520  \\

\textbf{Ours}                      &  -   &  \textbf{0.074} \\

  \end{tabular}
  }}
  \vspace{4pt}
  \caption{\label{tab:matterport_inference_speed} \textbf{Single image inference speed comparison:}
  Our image-to-image network is significantly faster than alternative off-the-shelf 3D geometry estimation methods. }
\end{table}

\begin{table}
  \centering
  {
  \footnotesize
  \begin{tabular}{lcc}
  & \textbf{Failed paths} & \textbf{Collisions} \\
  \shline

SSCNet \cite{song2017semantic}                &  0.643   &  0.207 \\
Convex hull                                                  &  0.608   &  0.180 \\
Bounding box \cite{qi2019deep} (Scannet)            &  0.569   &  0.157 \\
Bounding box \cite{qi2019deep} (Sun RGBD)           &  0.575   &  0.162 \\
Predicted visible ground                                  &  0.512   &  0.126 \\
Nothing traversable  ($S^* = 0$)                             &  0.616   &  0.198 \\
\textbf{Ours}                                                         &  \textbf{0.498}   &  \textbf{0.109} \\

\hline
Ground truth                                                 &  0.255   &  0.040 \\
  \end{tabular}
  }
  \vspace{3pt}
  \caption{\label{tab:matterport_path_planning} \textbf{Path planning evaluation on the Matterport dataset:} `Collisions' averages the total fraction of each path spent in space marked as non-traversable  by the ground truth, while a path is `failed' if it leaves ground-truth traversable space at any single point. Lower scores are better in both columns.}
\end{table}

\section{Use case: Path Planning}\label{Sec:PathPlanning}

One important use case for our system is to assist in the planning of paths, \eg for an augmented reality character.
For each Matterport test image we choose a random pixel on the ground truth `visible ground' mask as the start point and a pixel in the ground truth `hidden ground' mask as the end point.
We plan a path between the two with  A* \cite{hart1968formal}, where the cost of traversing pixel $j$ is $1 - s^*_{j}$, where $s^*_j$ is the unthresholded sigmoid output.
A planned path is  `failed' if it leaves the ground truth traversable area at any point; we also count the fraction of pixels in each path which leave the ground truth traversable area as `collisions'.
Results are shown in \mbox{Table~\ref{tab:matterport_path_planning}}, and examples of planned paths are shown in \mbox{Figure~\ref{fig:teaser}} and in the supplementary material.

\section{Conclusions}
In this work we have presented a novel representation for predicting scene geometry beyond the line of sight, and we have shown how to learn to predict this geometry using only stereo or depth-camera video as input.
We demonstrated our system's performance on a range of challenging datasets, and compared against several strong baselines.
Future work could address temporal consistency or persistent predictions.

\vspace{8pt}
{
\small
\paragraph{Acknowledgements}
Thanks to Eugene Valassakis for his help preparing this work's precursor \cite{watson2018dissertation}.
Special thanks also to Galen Han and Daniyar Turmukhambetov for help capturing, calibrating and preprocessing our handheld camera footage, and to Kjell Bronder for facilitating the dataset annotation.
}

\clearpage

\bibliographystyle{ieee}
{\small
\bibliography{main}
}

\end{document}

%% file: figs/kitti_results/index.tex
\begin{tabular}{cccccc}
\includegraphics[height=2cm]{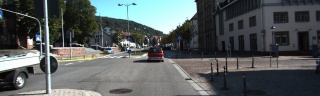} & 
\includegraphics[height=2cm]{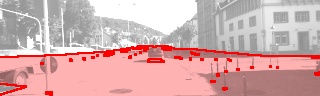} & 
\includegraphics[height=2cm]{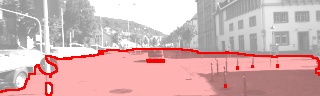} & 
\includegraphics[height=2cm]{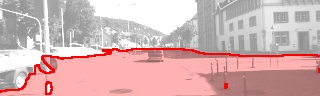} & 
\includegraphics[height=2cm]{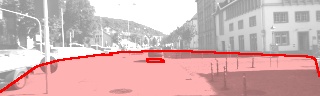} & 
\\
\includegraphics[height=2cm]{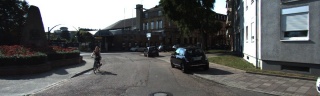} & 
\includegraphics[height=2cm]{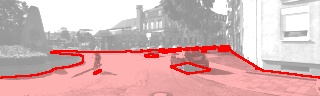} & 
\includegraphics[height=2cm]{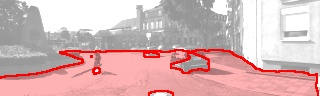} & 
\includegraphics[height=2cm]{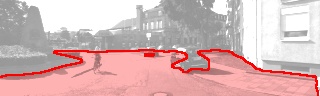} & 
\includegraphics[height=2cm]{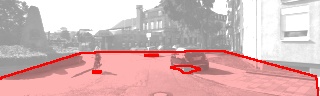} & 
\\
\includegraphics[height=2cm]{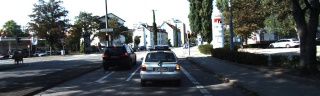} & 
\includegraphics[height=2cm]{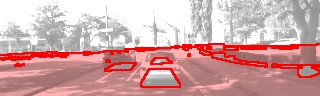} & 
\includegraphics[height=2cm]{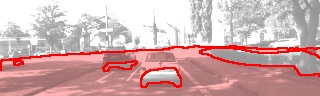} & 
\includegraphics[height=2cm]{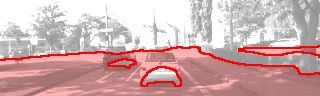} & 
\includegraphics[height=2cm]{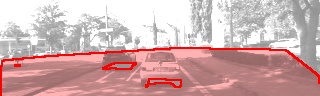} & 
\\
\includegraphics[height=2cm]{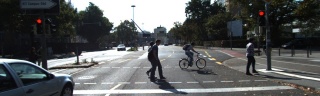} & 
\includegraphics[height=2cm]{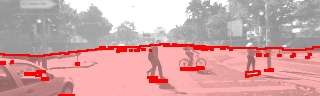} & 
\includegraphics[height=2cm]{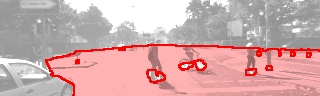} & 
\includegraphics[height=2cm]{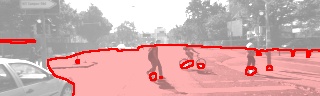} & 
\includegraphics[height=2cm]{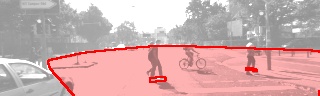} & 
\\
\includegraphics[height=2cm]{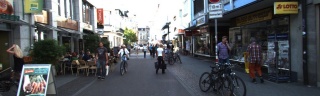} & 
\includegraphics[height=2cm]{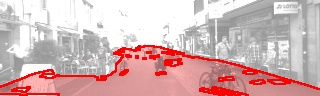} & 
\includegraphics[height=2cm]{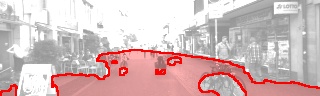} & 
\includegraphics[height=2cm]{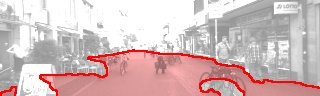} & 
\includegraphics[height=2cm]{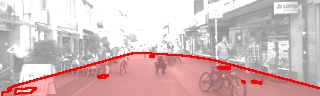} & 
\\
\includegraphics[height=2cm]{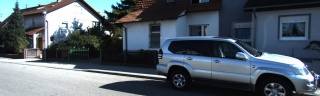} & 
\includegraphics[height=2cm]{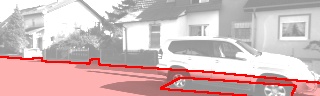} & 
\includegraphics[height=2cm]{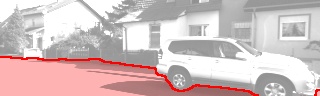} & 
\includegraphics[height=2cm]{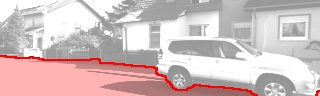} & 
\includegraphics[height=2cm]{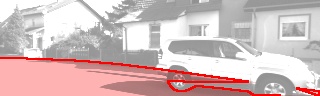} & 
\\ \vspace{3pt}
\LARGE Input & 
\LARGE Ground truth & 
\LARGE Ours & 
\LARGE Ours no depth mask & 
\LARGE Bounding box \cite{mousavian20173d} & 
\\
\end{tabular}

%% file: figs/matterport_results/index.tex
\setlength\tabcolsep{1.5pt} 
\begin{tabular}{ccccc}
\raisebox{20pt}{\rotatebox{90}{Input}} & 
\includegraphics[height=2cm]{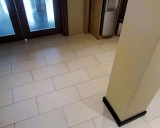} & 
\includegraphics[height=2cm]{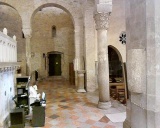} & 
\includegraphics[height=2cm]{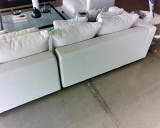} & 
\includegraphics[height=2cm]{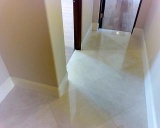} \\

\raisebox{4pt}{\rotatebox{90}{Ground truth}} & 
\includegraphics[height=2cm]{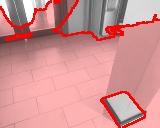} & 
\includegraphics[height=2cm]{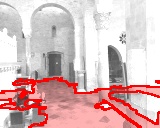} & 
\includegraphics[height=2cm]{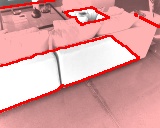} & 
\includegraphics[height=2cm]{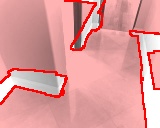} \\

\raisebox{20pt}{\rotatebox{90}{Ours}} & 
\includegraphics[height=2cm]{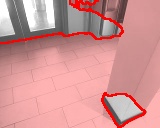} & 
\includegraphics[height=2cm]{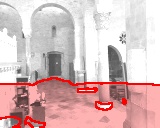} & 
\includegraphics[height=2cm]{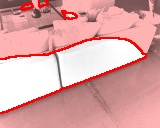} & 
\includegraphics[height=2cm]{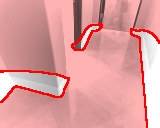} \\

\raisebox{5pt}{\rotatebox{90}{B.~box \cite{qi2019deep} }} & 
\includegraphics[height=2cm]{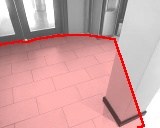} & 
\includegraphics[height=2cm]{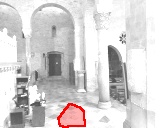} & 
\includegraphics[height=2cm]{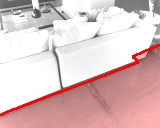} & 
\includegraphics[height=2cm]{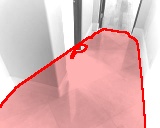} \\

\raisebox{10pt}{\rotatebox{90}{SSCNet+}} & 
\includegraphics[height=2cm]{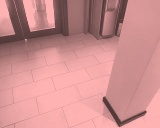} & 
\includegraphics[height=2cm]{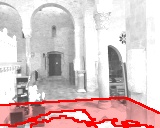} & 
&
\includegraphics[height=2cm]{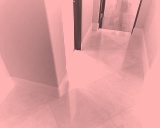} 
\\
\\
\end{tabular}